\documentclass[runningheads]{llncs}
\usepackage{graphicx}
\usepackage[table]{xcolor}
\definecolor{Graylight}{gray}{0.9}
\usepackage{tikz}
\usepackage{comment}
\usepackage{amsmath,amssymb} % define this before the line numbering.
\usepackage{color}
\usepackage{bm}
\usepackage[pagebackref,breaklinks,colorlinks]{hyperref}

\usepackage{graphicx}
\usepackage{amsmath}
\usepackage{amssymb}
\usepackage{booktabs}
\usepackage{multirow}
\usepackage{caption}
\usepackage{subcaption}

\newcommand{\tablestyle}[2]{\setlength{\tabcolsep}{#1}\renewcommand{\arraystretch}{#2}\centering\footnotesize}

\usepackage[accsupp]{axessibility}  % Improves PDF readability for those with disabilities.

\begin{document}
% \linenumbers
\pagestyle{headings}
\mainmatter
\def\ECCVSubNumber{1203}  % Insert your submission number here

% \title{Latent MAE: An Efficient Framework for Masked Image Modeling} % Replace with your title
\title{Bootstrapped Masked Autoencoders\\for Vision BERT Pretraining}

%\begin{comment}
\titlerunning{BootMAE for Vision BERT Pretraining}
\author{
Xiaoyi Dong$^{1}$\thanks{Work done during an internship at Microsoft Research Asia. $\dagger$ Dongdong Chen is the corresponding author.},  Jianmin Bao$^{2}$, Ting Zhang$^{2}$, Dongdong Chen$^{3\dagger}$,\ Weiming Zhang$^{1}$, Lu Yuan$^{3}$, Dong Chen$^{2}$, Fang Wen$^{2}$,  Nenghai Yu$^{1}$  \\
$^{1}$University of Science and Technology of China  \\
$^{2}$Microsoft Research Asia
$^{3}$Microsoft Cloud $+$ AI \\
{\tt\small\{dlight@mail., zhangwm@, ynh@\}.ustc.edu.cn } 
{\tt\small cddlyf@gmail.com }\\
{\tt\small\{jianbao, Ting.Zhang, luyuan, doch, fangwen \}@microsoft.com } 
}
\institute{}
\authorrunning{X Dong et al.}
%\end{comment}
%******************
\maketitle

\begin{abstract}
We propose bootstrapped masked autoencoders (BootMAE), a new approach for vision BERT pretraining. BootMAE improves the original masked autoencoders (MAE) with two core designs: 1) momentum encoder that provides online feature as extra BERT prediction targets; 2) target-aware decoder that tries to reduce the pressure on the encoder to memorize target-specific information in BERT pretraining. The first design is motivated by the observation that using a pretrained MAE to extract the features as the BERT prediction target for masked tokens can achieve better pretraining performance. Therefore, we add a momentum encoder in parallel with the original MAE encoder, which bootstraps the pretraining performance by using its own representation as the BERT prediction target. In the second design, we introduce target-specific information (e.g., pixel values of unmasked patches) from the encoder directly to the decoder to reduce the pressure on the encoder of memorizing the target-specific information. Thus, the encoder focuses on semantic modeling, which is the goal of BERT pretraining, and does not need to waste its capacity in memorizing the information of unmasked tokens related to the prediction target. Through extensive experiments, our BootMAE achieves $84.2\%$ Top-1 accuracy on ImageNet-1K with ViT-B backbone, outperforming MAE by $+0.8\%$ under the same pre-training epochs. BootMAE also gets $+1.0$ mIoU improvements on semantic segmentation on ADE20K and $+1.3$ box AP, $+1.4$ mask AP improvement on object detection and segmentation on COCO dataset. Code is released at \url{https://github.com/LightDXY/BootMAE}.

\end{abstract}

\vspace{-1mm}
\section{Introduction}
\vspace{-1mm}
Self-supervised representation learning~\cite{xie2016unsupervised,van2018representation,yang2016joint,zhuang2019local,ermolov2021whitening,henaff2020data,chen2020simple}, aiming to learn transferable representation from unlabeled data, has been a longstanding problem in the area of computer vision. Recent progress has demonstrated that large-scale self-supervised representation learning leads to significant improvements over the supervised learning counterpart on challenging datasets.
Particularly,
masked image modeling (MIM) in self-supervised pre-training for vision transformers has shown remarkably impressive downstream performance in a wide variety of computer vision tasks~\cite{dosovitskiy2020image,bao2021beit}, attracting increasing attention.

MIM aims to recover the masked region based on remaining visible patches. Essentially, it learns the transferable representation through modeling the image structure itself by content prediction. A very recent work, masked autoencoder (MAE)~\cite{he2021masked}, 
introduces an asymmetric encoder-decoder structure where the encoder only operates on visible patches, and the output representation of the encoder along with masked tokens are fed into a lightweight decoder.
Shifting the mask tokens into the small decoder results in a large reduction in computation. Besides efficiency, it also
achieves competitive accuracy (87.8\%), equipped with the ViT-Huge backbone, among methods that only use ImageNet-1K data.

In this paper, we introduce bootstrapped masked autoencoders (BootMAE), a new framework for self-supervised representation learning with two core designs. Firstly, we observe that with the same structure design as MAE, just changing the MIM prediction target from the pixels to the representation of a pretrained MAE encoder boosts the ImageNet classification accuracy from $83.4\%$ to $83.8\%$ using a ViT-Base backbone.
Motivated by this observation, we propose to use a momentum encoder to provide an extra prediction target. The momentum encoder is a temporal ensemble of the MAE encoder, \emph{i.e.}, the weights are parameterized by an exponential moving average (EMA) of the MAE encoder parameters~\cite{he2020momentum,grill2020bootstrap}. For each iteration, we pass the full image to the momentum encoder to provide ground-truth representation for masked patches, and pass the masked image to the encoder followed by a predictor to generate predictions for masked patches.
We hypothesize that as the training proceeds, the momentum encoder provides dynamically deeper semantics than fixed targets via bootstrapping. We keep the pixel regression branch in MAE as a good regularization in differentiating images. Moreover, it also provides guidance for the model to learn reasoning about low-level textures. Such multiple supervision helps learn the representation that benefits broader tasks including high-level recognition and accurate pixel-wise prediction that requires low-level information.

Secondly, we propose the target-aware decoder that tries to reduce the pressure on the
encoder to memorize target-specific information and encourage the encoder focus on semantic modeling that benefits for pre-training.
Recall that MIM aims to recover the missing region given the visible patches. It is based on the fact that natural images, regardless of their diversity, are highly structured (for example, the regular pattern of buildings, the structured shape of cars). The goal of MIM is to enable the model to understand this structure, or so-called semantics, or equivalently the relationship of different patches in the prediction target space (either pixel space or feature space).
Afterwards, the predictions are made by two indispensable ingredients: the knowledge of this structure and the target-specific information (\emph{e.g.}, pixel values) of the visible patches.
Yet previous MIM methods couple the two ingredients in a single module, wasting the model capability in ``memorizing" the target-specific information of visible patches.
In comparison, we try to decouple them so that the encoder exploits its whole model capability for structure learning. More specifically, the target-specific information is explicitly and continuously given to the decoder, just like we humans always see the visible patches when making visual predictions.

\begin{figure}[t]\centering
\includegraphics[width=0.995\linewidth]{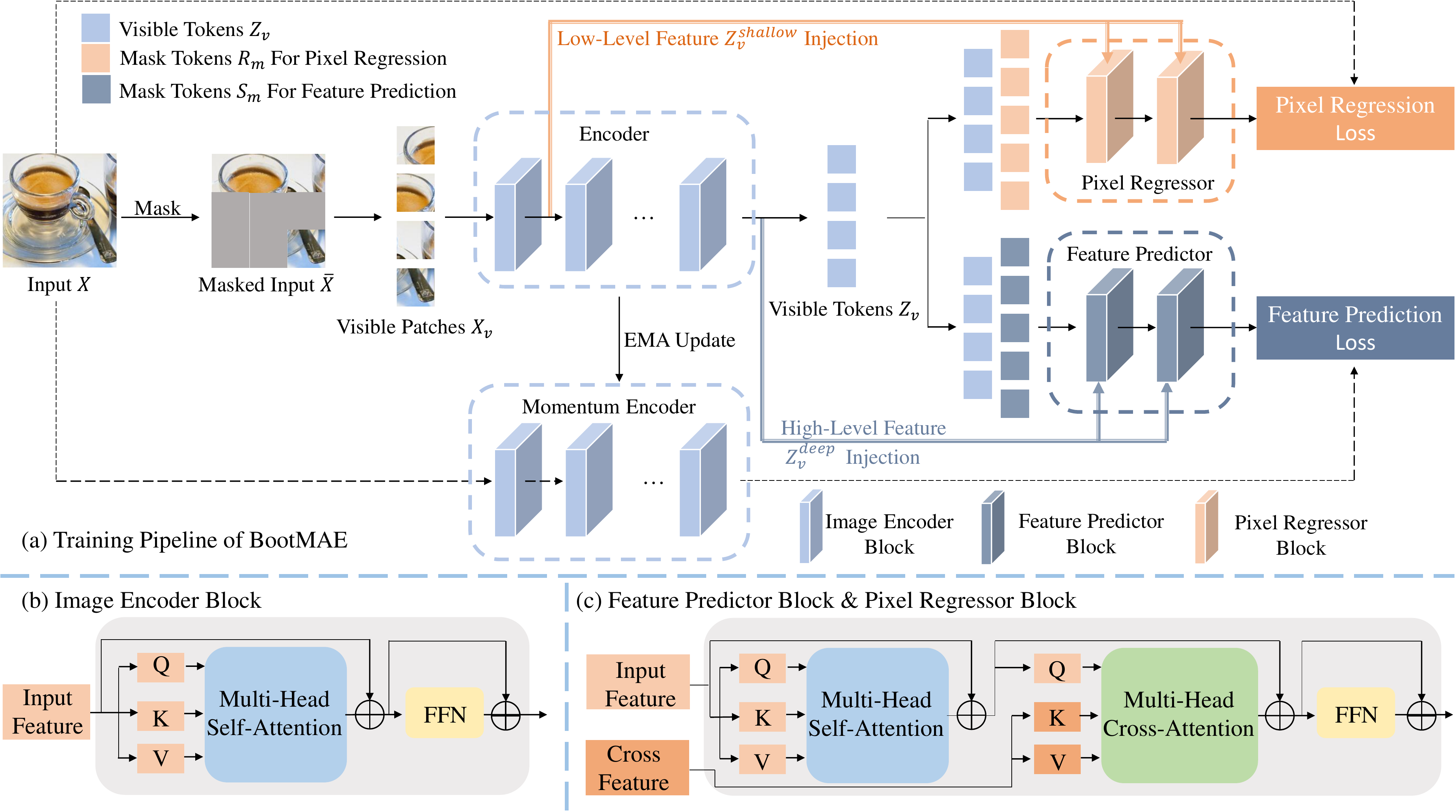}\vspace{-.7em}
\caption{Illustrating the details of our BootMAE in (a) the overall framework and training pipeline, (b) the image encoder block, (c) the feature predictor block \& pixel regressor block.}
\label{fig:pipeline}
\vspace{-4mm}
\end{figure}

In summary,
our framework, as illustrated in Figure~\ref{fig:pipeline}, contains four components: (1) an encoder that aims to capture the structure knowledge; (2) a regressor that takes the structure knowledge from the encoder along with the low-level context information for pixel-level regression; (3) a predictor that takes the structure knowledge from the encoder and the high-level context information for latent representation prediction; (4) feature injection modules in both regressor decoder and predictor decoder, responsible for incorporating each own necessary target-specific information.

In addition, we find that masking strategy is crucial for these two different prediction targets. They favor different masking strategies. Particularly, pixel regression relies on random masking while block-wise masking is better for feature prediction.
The reason might be that block-wise masking tends to remove large blocks, which is a difficult task
for pixel regression as pixel regression heavily relies on hints from local neighbors for prediction.
While for feature prediction not compelled by precise pixel-wise alignment, a large masked patch is more helpful for the model to reason about the semantic structure.

In the experiments we demonstrate the effectiveness of our framework in various downstream tasks including image classification, object detection and semantic segmentation. Our approach achieves superior performance than previous supervised methods as well as self-supervised methods. We also provide extensive ablation studies validating that the two core designs in our model works. 
We further provide comprehensive comparison with MAE in various epochs and various models and show our framework achieves consistently better performance.

\vspace{-1mm}
\section{Related Works}
\vspace{-1mm}
Computer vision has made tremendous progress on image content understanding in the past decade. The features learned by neural networks trained on ImageNet using over 1 million images associated with labels usually generalize very well across tasks~\cite{donahue2014decaf,sun2017revisiting,kolesnikov2020big,carreira2017quo}. Another line of image content understanding explores whether such semantically informative features can be learned through raw images alone~\cite{dosovitskiy2014discriminative,doersch2015unsupervised,xie2016unsupervised,van2018representation,yang2016joint,zhuang2019local,ermolov2021whitening,henaff2020data,chen2020simple,he2020momentum}. Representative methods along this line include autoencoding, clustering based, contrastive learning and masked image modeling.

\noindent \textbf{Autoencoding.}
Autoencoding (AE)~\cite{hinton2006reducing,bengio2009learning} is a type of neural networks used to learn a representation (embedding) for unlabeled data. 
It consists of an encoder that maps data to a low-dimensional latent embedding and a decoder that recovers the data from the latent embedding, 
with the goal of learning a compact feature representation for the data.
AE is commonly used for feature selection and feature extraction.
The denoising autoencoder (DAE)~\cite{vincent2008extracting} learns the representation robust to noise as the observed data in encoder is an addition of the original data and the noise. The decoder aims to undo the noise and recover the original data. 
Numerous efforts generalize DAE using different noise modelings, such as masking pixels~\cite{vincent2010stacked,pathak2016context,chen2020generative}, removing color channels~\cite{zhang2016colorful,larsson2016learning}, and shuffling image patches~\cite{noroozi2016unsupervised} and so on.

\noindent \textbf{Clustering based methods.}
Clustering is a class of unsupervised learning methods that has been widely studied in the computer vision community. Traditional works are mostly designed under the assumption that feature representation is fixed.
With the emerging of deep learning area, lots of efforts~\cite{yang2016joint,xie2016unsupervised} explore adapting clustering to the end-to-end training to jointly learn the feature representation as well as clustering. The representative work DeepCluster~\cite{caron2018deep} utilizes k-means to generate pseudo-labels to alternatively update the weights of the convnets and the clustering assignments of the image descriptors. Recently, several attempts~\cite{asano2019self,caron2020swav} aim to maximize the mutual information between the pseudo labels and the input data, scaling up to large datasets.

\noindent \textbf{Contrastive learning.}
Contrastive learning aims to learn an embedding space where similar data pairs stay close to each other while dissimilar data pairs are far apart.
In the self-supervised scenario, 
it can be interpreted as a special case of clustering where each instance itself forms a class. Thus, the positive pairs are formed by two augmented views of the same image and the negative paris are views from different images.
The typical methods include MOCO~\cite{he2020momentum,chen2020improved,chen2021empirical}, SimCLR~\cite{chen2020simple,chen2020big}, BYOL~\cite{grill2020bootstrap} and more~\cite{oord2018representation,li2021improve,bachman2019learning}.
However, contrastive-based methods rely heavily on the data augmentation strategies that need to introduce non-essential variations while without modifying semantic meanings.
Crucial augmentations include random cropping and random color distortion.
Meanwhile, large quantities of negative samples are usually required in order to avoid trivial solution in which the model outputs a constant representation for all data.

\noindent \textbf{Masked image modeling.}
Masked image modeling for self-supervised pre-training has recently grown in popularity motivated by the success of BERT pre-training in NLP~\cite{devlin2018bert}. ViT~\cite{dosovitskiy2020image} and BEiT~\cite{bao2021beit} are two initiatives along this direction.
MIM that predicts masked patches from visible ones in a sense can be viewed as context prediction.
Feature representation learned through such within-image context prediction shows surprisingly strong performance in downstream tasks. 
Recently, lots of works~\cite{he2021masked,fang2022corrupted} exploring MIM have been concurrently developed from different perspectives.
The efforts include (i) framework design, such as MAE~\cite{he2021masked}, SplitMask~\cite{el2021large}, SimMIM~\cite{xie2021simmim}, CAE~\cite{chen2022context};
(ii) prediction targets, such as PeCo~\cite{dong2021peco}, MaskFeat~\cite{wei2021masked}, data2vec~\cite{baevski2022data2vec}, iBOT~\cite{zhou2021ibot}; (iii) video extension BEVT \cite{wang2022bevt}; (iv) integration with vision-language contrastive learning FaRL\cite{zheng2022general}.
Our work belongs to the first group and introduces a novel framework called Bootstrapped MAE.
We progressively bootstrap the latent representation in MAE to learn dynamically deeper semantics. Moreover, in comparison with previous methods coupling the context information with semantic modeling in a single model, we separate them by explicitly passing the context information to the decoder so that the encoder leverage the whole model for structure learning. 

\vspace{-1mm}
\section{Approach}
\vspace{-1mm}
In this section, we introduce our Bootstrapped MAE framework in details.
As illustrated in Figure~\ref{fig:pipeline}, our framework contains four components: 1)  the encoder network focusing on learning the structure knowledge; 2) the pixel regressor decoder network aiming to predict the missing pixels of the masked region given the structure knowledge from the encoder and the context information from the visible patches,i.e.,  pixel values or low-level features in this case; 3) the feature predictor decoder network attempting to make feature predictions for the masked region given the same structure knowledge from the encoder and the context information of the visible patches, i.e., high-level feature information in this case;
and 4) feature injection modules that feed each own context information into the regressor and the predictor explicitly and continuously in each decoder layer. After self-supervised pre-training with our BootMAE, we adopt the encoder network for various downstream tasks.

Formally, suppose an input image is $\bm{X} \in \mathbb{R}^{H\times W \times C}$, where $H$ and $W$ denote the image height and image width and $C$ denotes the channel number, we first split it into non-overlapping patches. This results in $N = H\times W / P^2$ patches with $P$ denotes the resolution of each patch. In this way, an image is represented by a number of patches $\bm{X} = \{x^1, x^2, \cdots, x^N\}$ with $x^n \in \mathbb{R}^{P^2C}$ denotes a vector reshaped from the image patch. 
Thereafter, a large fraction of, say $N_m$ patches are randomly sampled to be masked and leave the remaining $N_v$ patches to be visible, $N = N_m +N_v$.
Let $\mathcal{M}$ be the index set of masked patches, $\bm{X}_v = \{x^k | k \notin \mathcal{M}\}$ denotes the set of visible patches and $\bm{X}_m = \{x^k | k \in \mathcal{M}\}$ denotes the set of masked patches, we have $\bm{X} = \bm{X}_v \cup \bm{X}_m$ and $\bm{X}_v \cap \bm{X}_m = \emptyset$.
Generally, each patch is associated with a positional embedding indicating the location of each patch. Therefore similarly, we have $\bm{P}_v$, positional embeddings for the visible patches and $\bm{P}_m$, positional embeddings for the masked patches.

\vspace{-1mm}
\subsection{Encoder}
\vspace{-1mm}
The encoder aims to exploit the whole capability to output a latent representation that models the image structure. 
Inspired by MAE~\cite{he2021masked},
the encoder only handles the visible patches $\bm{X}_v$ for training efficiency and outputs the latent representation $\bm{Z}_v$. 
Specifically, we first project each visible patch into an embedding and add a positional embedding on each embedding to ensure the awareness of position for each patch. After this, the combined embedding is fed to a ViT~\cite{dosovitskiy2020image} composed of a stack of standard vision Transformer blocks based on self-attention. That is,
\begin{equation}
	\bm{Z}_v = \text{Enc}(\bm{X}_v,\bm{P}_v).
\end{equation}
The computation and memory are really efficient even for large scale models as only a small subset (e.g.,25\%) of the image patches needs to be handled by the encoder.
Moreover, the elimination of the special mask tokens bridges the gap between pre-training and fine-tuning as the fine-tuning stage sees real visible patches without any mask token~\cite{he2021masked,fang2022corrupted}.
The mask region (e.g., 75\%) is randomly sampled from the image.

We find that the masking strategy is crucial as different prediction targets favor different masking strategies.
We study this masking strategy and provide more analysis in the experiments.
We provide explanations about which masking strategy is suitable for which prediction target, and reach a conclusion offering guidance about the choice of masking strategy.
In our implementation, we adopt the block-wise masking strategy.
The output is further normalized to $\hat{\bm{Z}}_v = \text{norm}(\bm{Z}_v)$ which captures the image structure and is fed to the subsequent decoders.

\vspace{-1mm}
\subsection{Feature injection module}
\vspace{-1mm}
As mentioned in the introduction, there are two indispensable ingredients for the decoder to make predictions: the structure knowledge and the corresponding context information from the visible patches.
Our feature injection module is designed to directly feed the context information into each layer of the decoder.
We argue that in this way, the encoder exploits the whole model capability to learn structure knowledge without considering ``memorizing" the context information of visible patches that is related to prediction target.

In particular, different prediction targets require different context information. Specifically, pixel-level prediction focusing on low-level details probably favors low-level context information of the visible patches while feature-level prediction attempting to predict semantic feature representation probably needs high-level context information of the visible patches.
Therefore, we feed the features from the shallow layer of the encoder to the regressor decoder and the features from the deep layer of the encoder to the predictor decoder. 
We use $\bm{Z}_v^{shallow}$ to represent the shallow features and $\bm{Z}_v^{deep}$ to represent the deep features of the encoder.

Instead of using addition or concatenation, we adopt a very elegant operator cross-attention.
To be specific, we leverage the feature from the encoder as keys and values and the features from the regressor/predictor as queries to perform cross-attention. This operator helps leverage the low-level information for better pixel reconstruction and the high-level semantics for feature prediction. We apply this cross attention right after the self-attention in each transformer block of the regressor and predictor.

\vspace{-1mm}
\subsection{Regressor}
\vspace{-1mm}
The regressor aims to recover the missing pixels as in~\cite{he2021masked}.
The pixel-level regression not only helps prevent the model from collapsing but also guide the model to learn reasoning about low-level textures.
The input of the regressor includes (1) the normalized latent representation output from the encoder and (2) the shallow features providing context information.
We add mask tokens $\bm{R}_m$ containing $N_m$ learnable vectors representing masked patches to be predicted. To ensure that each mask token is aware of its location in the image, we add the positional embedding to each mask token. We adopt a lightweight architecture for the regressor, consisting of two vision transformer blocks and 
a fully-connected layer to predict missing pixels. Let the output of the regressor be $\bar{\bm{X}}$, the formulation can be written as,
\begin{equation}
	\bar{\bm{X}} = \text{Reg}(\hat{\bm{Z}}_v, \bm{Z}_v^{shallow}, \bm{R}_m, \bm{P}_m).
\end{equation}
The regressor makes prediction based on the structure knowledge in $\hat{\bm{Z}}_v$ and the context information in $\bm{Z}_v^{shallow}$.

\vspace{-1mm}
\subsection{Predictor}
\vspace{-1mm}
The predictor aims to predict the feature representation of the masked patches.
This high-level feature prediction target guides the model to learn reasoning about high-level semantics. Moreover, the prediction groundtruth is the representation itself which evolves along with the training, providing richer and deeper semantics than fixed targets.
The input of the predictor includes (1) the normalized latent representation same with the regressor and (2) the deep features providing context information different from the regressor.
We also add another set of mask tokens $\bm{S}_m$ representing the masked patches to be predicted and associate them with positional embeddings.
The predictor decoder network consists of two transformer blocks with a MLP layer for prediction. 
Say the output of the predictor is $\bar{\bm{F}}$, the formulation can be written as,
\begin{equation}
	\bar{\bm{F}} = \text{Pre}(\hat{\bm{Z}}_v, \bm{Z}_v^{deep}, \bm{S}_m, \bm{P}_m).
\end{equation}
Similarly, the predictor makes prediction based on the structure knowledge in $\hat{\bm{Z}}_v$ and the context information in $\bm{Z}_v^{deep}$.

\vspace{-1mm}
\subsection{Objective function}
\vspace{-1mm}
The regressor and the predictor output all predictions for both visible patches as well as masked patches, but only the predictions for masked patches are involved in the loss calculation. 
For the regressor, each element in the output is a vector of pixel values representing a patch. We use normalized pixels as the reconstruction target for groundtruth as MAE~\cite{he2021masked}. The objective function for the regressor is,
\begin{equation}
\mathcal{L}_{R} = \sum_{k \in \mathcal{M}} \frac 1 {P^2C}||g_m^{k} - \bar{x}_m^k||_2^2,
\end{equation}
where $g_m^{k}$ is the normalized patch of $x_m^{k}$ using the mean and standard deviation computed from all pixels in that patch. $\bar{x}_m^k$ is the reconstructed masked patch in $\bar{\bm{X}}$

For the predictor, the prediction feature groundtruth is the latent representation itself by passing a full image into the momentum encoder where the weights are parameterized by an exponential moving average of the MAE encoder. Let $\bm{F} = \text{Enc}_{ema}(\bm{X},\bm{P})$ be the groundtruth, the objective function over the masked patches for predictor is,
\begin{equation}
 \mathcal{L}_{P} = \sum_{k\in \mathcal{M}} \frac 1 {\text{\#dim}} ||f_m^{k} - \bar{f}_m^k||_2^2,   
\end{equation}
where \#dim denotes the feature dimension of the token, and $f,\bar{f}$ is one of token in $F,\bar{F}$. The overall loss is a weighted sum,
\begin{equation}
 \mathcal{L} = \mathcal{L}_{R} + \lambda \mathcal{L}_{P},
\end{equation}
where $\lambda$ is the hyperparameter tuning the loss weight and set to 1 by default.

\vspace{-1mm}
\section{Experiments}
\vspace{-1mm}

\subsection{Implementations}
\vspace{-1mm}
We experiment with the standard ViT\footnote{we didn't use some recent techniques like relative position or layer scaling.} base and large architectures,
ViT-B (12 transformer blocks with dimension 768) and ViT-L (24 transformer blocks with dimension 1024) for the encoder. 
The regressor and the predictor consist of 2 transformer blocks as mentioned above. The dimension of the regressor is set to 512 while the the dimension of the predictor is set to the same as the encoder for feature prediction.
The input is partitioned $14\times 14$ patches from the image of $224\times 224$, and each patch is of size 16x16. Following the setting in MAE, we only use
standard random cropping and horizontal flipping for data augmentation.
We find that the different prediction tasks favor different masking strategies. We choose the block-wise masking strategy to benefit for feature prediction.
The total masking ratio is 75\%, same with that in MAE~\cite{he2021masked}.
Both ViT-B and ViT-L model are trained for 800 epochs with batch size set to 4096.
We use Adam~\cite{kingma2014adam} and a cosine schedule~\cite{loshchilov2016sgdr} with a single cycle where
we warm up the learning rate for 40 epochs to $2.4e^{-3}$. The learning rate is further annealed following the cosine schedule. 
Our proposed method is pre-trained on ImageNet. The regressor and the predictor are only used during pre-training. After pre-training, only the encoder is used to generate the image representation.

For ImageNet experiments, we average pool the output of the last transformer of the encoder and feed it to a softmax-normalized classifier.
We evaluate the pre-trained feature representation using end-to-end fine-tuning along with the backbone model. We fine-tune 100 epochs for ViT-B and 50 epochs for ViT-L.
The learning rate are warmed up to 0.005 for 20 epochs for ViT-B and 0.0015 for 5 epochs for ViT-L, after which followed by cosine schedule. The evaluation metric is top-1 validation accuracy of a single $224\times 224$ crop.

\vspace{-1mm}\subsection{Analysis of BootMAE}
\vspace{-1mm}

\begin{table}[t]
	\centering
	\footnotesize
	\caption{The effect of bootstrapped feature prediction. The performance with pre-trained 300 epochs gets improvement from 83.2\% to 83.6\%, achieving the same performance with the vanilla MAE with pre-trained 1600 epochs.}
	\label{tab:bootstrap}
	\setlength{\tabcolsep}{4mm}{
		\begin{tabular}{l|c|c }
			\hline
			Model & Pre-train Epoch & Fine-tuning \\% & Linear probing  \\
			\hline
			MAE & 1600 & 83.6  \\
			MAE & 800 & 83.4  \\
			MAE & 300 & 83.2  \\
			MAE w bootstrapped feature prediction &  300 & 83.6 \\
			\hline
	\end{tabular}}
	
	\vspace{-4mm}
\end{table}

\begin{table}[t]
	\centering
	\footnotesize
	\caption{Ablation studies showing the effect of feature injection module in our framework. Providing context for both regressor and predictor achieves the best performance, suggesting that in this target-aware decoder design, the encoder indeed learns stronger semantic modeling.}
	\label{tab:feature}
	\setlength{\tabcolsep}{3.5mm}{
		\begin{tabular}{l|c|c|c }
			\hline
			Model& Context for regressor & Context for predictor &  Fine-tuning    \\
			\hline
			BootMAE & $\times$ & $\times$ &  83.6  \\
			BootMAE & $\checkmark$ & $\times$ & 83.9 \\
			\rowcolor{Graylight} 
			BootMAE & $\checkmark$& $\checkmark$ & 84.0   \\
			\hline
	\end{tabular}}
	\vspace{-4mm}
\end{table}

\noindent \textbf{Bootstrapped feature prediction.}
One core design of our framework is the bootstrapped feature prediction that predicts the iteratively evolved latent representation of the image to enable the model to learn from dynamically richer semantic information.
Here we investigate the effect of adding this proposed bootstrapped feature prediction branch. The compared models are the vanilla MAE and the MAE with an additional bootstrapped feature prediction without the feature injection module. 
The comparison results are presented in Table~\ref{tab:bootstrap}. 
We observe that the performance with pre-trained 300 epochs gets improvement from 83.2\% to 83.6\%, achieving the same performance with the vanilla MAE with pre-trained 1600 epochs. This demonstrates the effectiveness of the proposed bootstrapped feature prediction. Based on this result, we further analysis BootMAE under the 300 epoch pre-training setting in the following.

\vspace{2mm}

\noindent \textbf{Feature injection.}
Another important design in our framework is the feature injection module, which provides different features representing different level of context information for the regressor and the predictor. Specifically, we explicitly feed the feature outputted from the first layer of the encoder to each layer of the regressor to ease the burden of the encoder in ``memorizing" the low-level details so that the encoder focuses on structure modeling.
Similarly, we directly feed the features from the last layer of the encoder to each layer of the predictor.
Here we study the effect of the proposed feature injection and the ablated results are shown in Table~\ref{tab:feature}. We can see that providing both regressor and predictor with each own necessary context achieves the best performance, the encoder indeed learns stronger semantic modeling due to the target-aware decoder design.

\begin{table}[t]
	\centering
	\footnotesize
	\caption{Results comparison of two different masking strategies, random masking and block-wise masking, with two different prediction targets, pixel-level target and feature-level target. This validates our hypothesis that pixel-level target favors random masking while feature-level target favors block-wise masking.}
	\label{tab:mask}
	\setlength{\tabcolsep}{5.5mm}{
		\begin{tabular}{l|cc|c }
			\hline
			\multirow{2}{*}{Prediction target}
			& \multicolumn{2}{c|}{Mask Strategy} &  \multirow{2}{*}{ Fine-tuning Accuracy} \\
			& Random & Block &\\
			\hline
			Pixel-level        & \checkmark  & & 83.2  \\
			Pixel-level        & &  \checkmark     & 82.8  \\
			Feature-level    & \checkmark  &  & 83.1 \\
			Feature-level    & &  \checkmark     & 83.6 \\
			\hline
	\end{tabular}}
	\vspace{-2mm}
\end{table}

\begin{figure}[t]\centering
\includegraphics[width=1\linewidth]{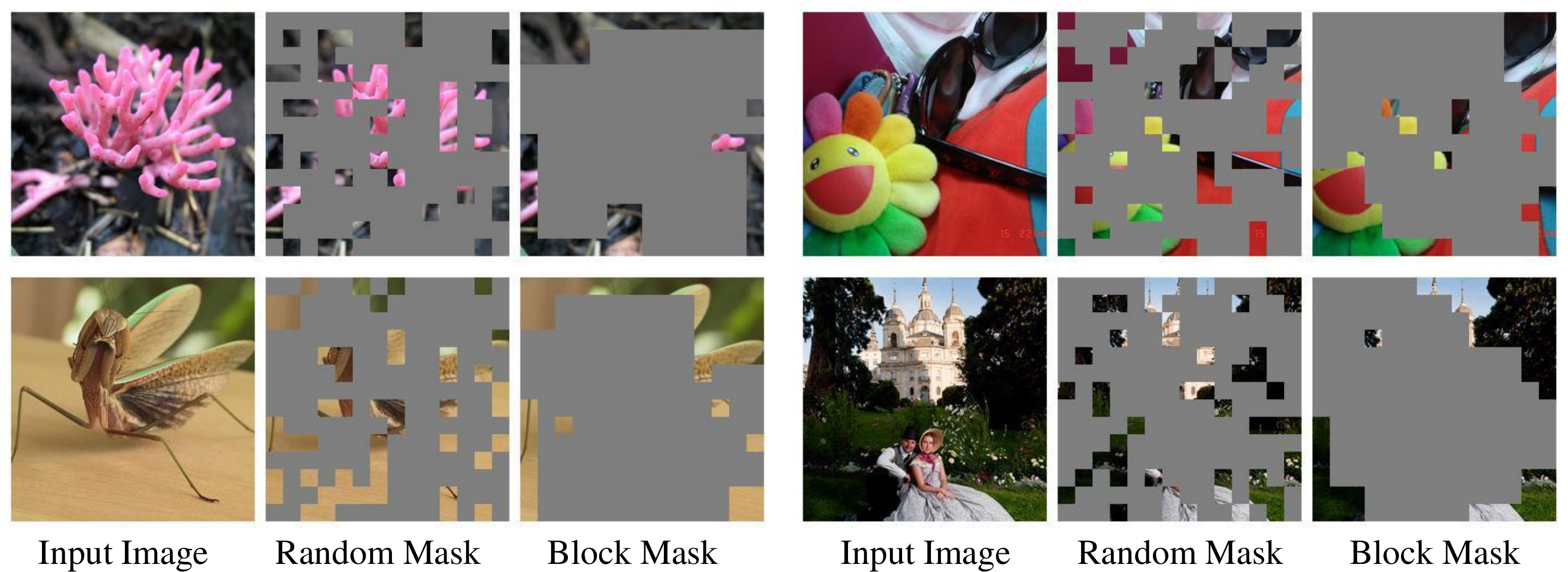}\vspace{-.7em}
\caption{Visualization of the two different masking strategies. The masked region is close to the visible region in random masking. For block-wise masking, a large continuous block is masked and most center patches are masked.}
\label{fig:masking}
\end{figure}

\begin{table*}[t]
	\centering
	\caption{We study the regressor and predictor design and ablate the performance in terms of (a) the regressor and predictor depth and (b) the feature dimension of the regressor.}
	\label{tab:decoder}
	\tablestyle{6pt}{1.05}
	\begin{subtable}[t]{0.5\linewidth}
		\resizebox{\linewidth}{!}{
			\setlength{\tabcolsep}{3.5mm}{
				\begin{tabular}{c|c}
					\hline
					\multicolumn{2}{c}{(a) The regressor and predictor depth}\\
					\hline
					Depth & Fine-tuning   \\ % & Linear probing 
					\hline
					1 & 83.4  \\
					\rowcolor{Graylight} 
					2 & 84.0   \\
					4 & 83.9   \\
					8 & 84.0   \\
					\hline
		\end{tabular}}}
	\end{subtable}\hfill
	\begin{subtable}[t]{0.5\linewidth}
		\resizebox{\linewidth}{!}{
			\setlength{\tabcolsep}{4mm}{
				\begin{tabular}{c|c}
					\hline
					\multicolumn{2}{c}{(b) The feature dimension of regressor}\\
					\hline
					\#Dim & Fine-tuning    \\ %& Linear  probing
					\hline
					256 & 83.8   \\
					384 & 84.0   \\
					\rowcolor{Graylight} 
					512 & 84.0   \\
					768 & 84.0   \\
					\hline
		\end{tabular}}}
	\end{subtable}
	
	\vspace{-4mm}
\end{table*}

\begin{table}[t]
	\centering
	\footnotesize
	\caption{We study the pre-training efficiency with the momentum encoder and report the performance as well as iteration time cost when feeding different fractions of the image to the momentum encoder.}
	\label{tab:inference}
	\setlength{\tabcolsep}{2mm}{
		\begin{tabular}{l|c|ccc}
			\hline
			\multirow{2}{*}{Model} & Fraction of Image & \multirow{2}{*}{Fine-tuning}  & Training  & \multirow{2}{*} {Speed Up}  \\
			& for Momentum Encoder & & Iter Time (s) &\\
			\hline
			MAE & ---  &  83.2 & 0.473 & $1\times$ \\
			BootMAE & 25\%   &  83.8  & 0.407 & $1.16\times$ \\
			
			BootMAE & 50\%   &  83.8  & 0.479 & $0.98\times$ \\
			
			BootMAE & 75\%   &  83.9  & 0.588 & $0.80\times$ \\
			\rowcolor{Graylight} 
			BootMAE & 100\% (Default)  & 84.0 & 0.660 & $0.72\times$  \\
		
			\hline
	\end{tabular}}
	
	\vspace{-4mm}
\end{table}

\vspace{2mm}
\noindent \textbf{Masking strategy.}
Then we study two widely used masking strategies in masked image modeling: random block-wise masking in~\cite{bao2021beit} and random masking in~\cite{he2021masked}. We show several examples of the two masking strategies in Figure~\ref{fig:masking}.
The masking ratio is the same and is set to 75\%. 
It has been observed in MAE~\cite{he2021masked} that block-wise masking degrades at such a large ratio for their model. While in other scenarios, we find that block-wise masking is better than random masking.
Here we provide explanations about why this is the case.

We suspect that the reason may come from the prediction target. Pixel-level prediction target pursuing precise pixel-wise alignment requires visible neighboring patches to provide texture information, thus favoring that the masked region should be close to the visible region.
While in block-wise masking, there always exists a larger continuous block of the image being masked and more masked patches are near the image center, making it difficult to the pixel-level prediction.
As for feature-level prediction which cares less about the textures/details, block-wise masking largely reduces the redundancy and most center patches are masked, forcing the model to learn reasoning about the semantics. 

We experiment the two masking strategies when using two different prediction targets and the results are given in Table~\ref{tab:mask}. Here we train MAE for 300 epochs with different prediction target: pixel (MAE default setting) or output feature of a 800 epoch pretrained MAE model.
The comparison validates our hypothesis analyzed above that pixel-level (feature-level) target favors random masking (block-wise masking). 
We adopt block-wise masking as bootstrapped feature prediction is key in our framework.

\vspace{2mm}
\noindent \textbf{Regressor and predictor design.}
Our regressor and predictor are pretty lightweight consisting of two transformer layers. In this section, we vary the network depth (number of Transformer blocks)
and experiment the performance when setting the depth to 1, 2, 4 and 8.
The results are reported in Table~\ref{tab:decoder} (a).
We can see that using depth 2 or 8 achieves the best fine-tuning performance while depth 2 enjoys more efficiency.
In addition we also study the feature dimension in regressor. Note that the feature dimension in predictor is set as the same with the encoder width. 
As shown in Table~\ref{tab:decoder} (b).
The fine-tuning accuracy with different dimensions is similar, except dim$=256$ which is too small.

\vspace{2mm}
\noindent \textbf{Pre-training efficiency with the momentum encoder.}
In our framework, we feed the full image to the momentum encoder to provide the feature prediction ground-truth. 
We observe that this extra inference incurs additional computation cost compared with MAE.
Here we present specific training iteration time in Table~\ref{tab:inference}.
The validation is conducted with A100 GPU and batch size 256 per GPU for all models.
We further study several variants that only a subset of the masked patches are fed into the momentum encoder and the prediction loss is only evaluated on this subset of masked patches. As the masking ratio is 75\%, we study three fractions: 75\% (all the masked patches), 50\% (sampled from masked patches), 25\% (also sampled from masked patches). We report the iteration time as well as the performance in Table~\ref{tab:inference}. We can see that as with a smaller fraction of patches to the momentum encoder, the iteration time cost gets fewer while the performance gets lower due to the model only learns from a fraction of the masked tokens.
It is worth noting that our method when feeding 25\% image patches to the momentum encoder achieves better performance than MAE while is more efficient.
This is because MAE adopt 8 layers for the decoder while our regressor and predictor only consist of 2 layers.

\begin{table}[t]
\centering
\small
\caption{Image classification accuracy (\%) comparison on ImageNet-$1$K of different methods using various backbones. -B, -L stands for using ViT-B, ViT-L model, respectively.  We report the fine-tuning and linear probing accuracy and our method BootMAE outperforms previous self-supervised methods. }
\label{tab:imagenet}
\resizebox{1\linewidth}{!}{
\setlength{\tabcolsep}{1mm}{
\begin{tabular}{lcccccc}
\toprule
\multirow{2}{*}{ Methods} & Pre-train  & Pre-train & \multicolumn{2}{c}{ViT-B} & \multicolumn{2}{c}{ViT-L} \\ & dataset & epochs & Fine-tuning & Linear & Fine-tuning & Linear \\

\midrule
\multicolumn{5}{l}{\textit{Training from scratch (i.e., random initialization)}} \\
ViT$_{384}$~\cite{dosovitskiy2020image}  & - & - & 77.9 & -  & 76.5 & - \\
DeiT~\cite{touvron2021training}        & - & - & 81.8 & - & -- & -  \\
ViT~\cite{he2021masked}           & - & - & 82.3 & - & 82.6 & -  \\
\midrule
\multicolumn{5}{l}{\textit{Self-Supervised Pre-Training on ImageNet-1K}} \\
DINO~\cite{caron2021emerging}      & IN-$1$K & 300 &  82.8 & 78.2 & -- & -- \\
MoCo v3~\cite{chen2021empirical}   & IN-$1$K & 300 &  83.2 & 76.7 & 84.1 & 77.6 \\
BEiT~\cite{bao2021beit}            & IN-$1$K + DALLE & 800 &  83.2 & 56.7 & 85.2 & 73.5 \\
MAE~\cite{he2021masked}     & IN-$1$K& 800 & 83.4 & 64.4 & 85.4 & 73.9 \\
MAE$^*$~\cite{he2021masked}     & IN-$1$K& 1600 & 83.6 & 68.0 & 85.9 & 76.6 \\
\rowcolor{Graylight} 
BootMAE    & IN-$1$K     & 300 &  84.0 & 64.1 & 85.4 &  74.8  \\
\rowcolor{Graylight} 
BootMAE    & IN-$1$K     & 800 & \underline{84.2} & 66.1 & \textbf{85.9} &  77.1  \\

\bottomrule
\end{tabular}}}

\vspace{-2mm}
\end{table}

\vspace{-1mm}
\subsection{ImageNet Classification Comparison}
\vspace{-1mm}
We compare our methods with previous state-of-the-art works on ImageNet-1K classification task. We report the top-1 validation accuracy for both fine-tuning and linear probing results in Table~\ref{tab:imagenet}. Compared to the supervised models trained from scratch, self-supervised pre-training methods achieve significant improvement, suggesting the effectiveness of pre-training. 

We further compare our framework with prior self-supervised pre-training models. We can see that the proposed BootMAE achieves the best fine-tuning performance either based on ViT Base network or based on ViT Large network.
For example, compared with the recent work MAE~\cite{he2021masked}, our bootMAE with ViT-B achieves 84.2\% top-1 accuracy with 0.8\% gain, and with ViT-L achieves 85.9\% with 0.5\% improvement.
We also report the linear probing accuracy. Our approach performs better than MIM based self-supervised methods, but not as good as the contrastive-based methods.
We suspect that contrastive learning methods pursue linear features by comparing across images while MIM based methods exploit within image structure.

\begin{figure}[t]\centering
\includegraphics[width=1\linewidth]{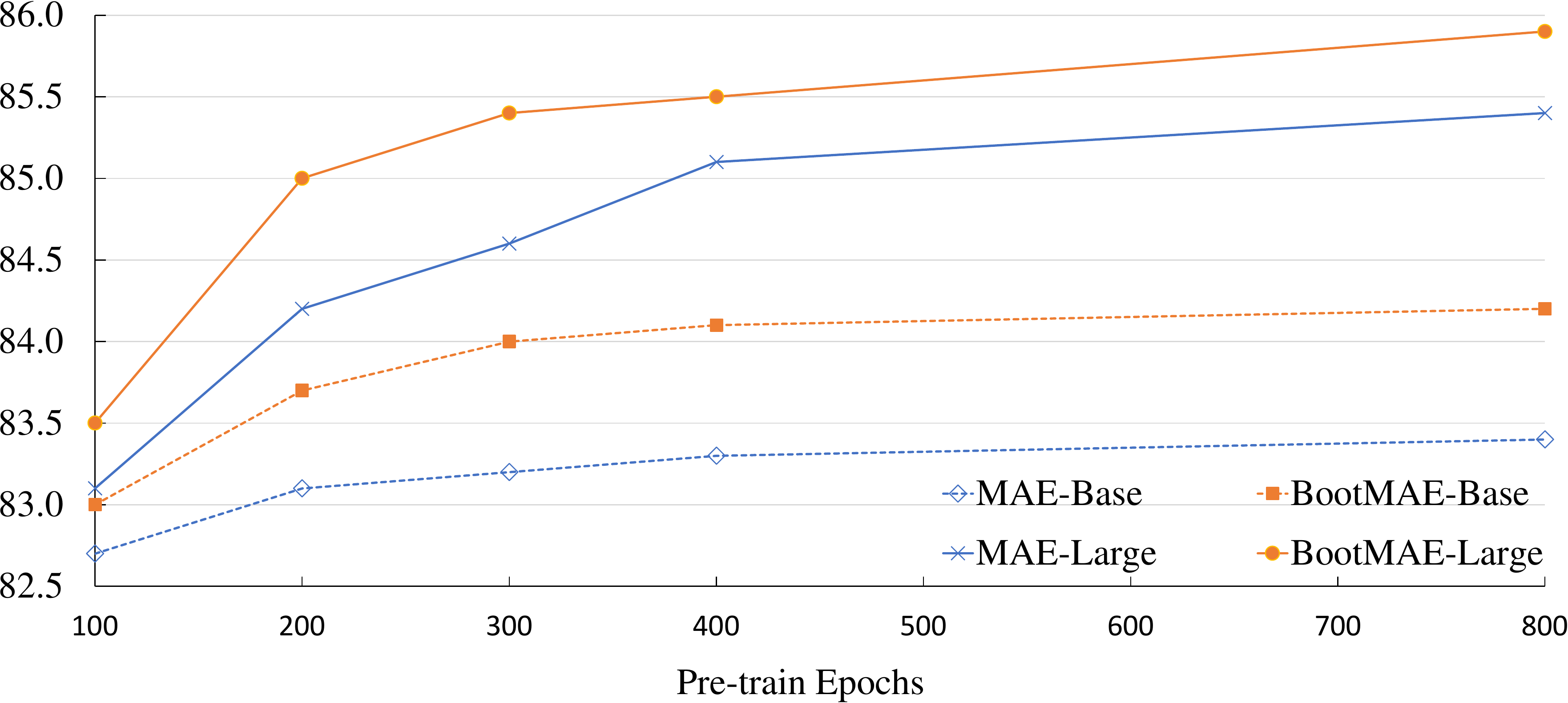}\vspace{-.7em}
\caption{Performance comparison with MAE in different pre-training epochs using ViT-B and ViT-L, showing that our BootMAE  consistently outperforms MAE.}
\label{fig:epoch}
\vspace{-4mm}
\end{figure}

In addition, we present comprehensive comparison with MAE under different pre-training epochs for both ViT-B and ViT-L. We plot the results in Figure~\ref{fig:epoch}. We can see that our approach consistently performs better than MAE. It is worth mentioning that the proposed bootMAE at 200 epochs achieves 83.7\% accuracy, which is alredy better than MAE pre-trained at 800 epochs.
This demonstrate that our approach is more efficient to achieve similar performance, though with the extra inference of the momentum encoder.
To be specific, under the same setting that using 16 V100 GPUs, MAE takes 51 hours for 800 epochs to get an 83.4\% accuracy, 
while our BootMAE only takes 18 hours for 200 epochs to get a better result 83.7\%.

\vspace{-1mm}
\subsection{Downstream Tasks}
\vspace{-1mm}
To further validate the learned visual representation of our BootMAE, we present transfer learning experiments on two popular downstream tasks.

\noindent \textbf{Semantic segmentation.}
We compare our model on the widely used semantic segmentation dataset ADE20K~\cite{zhou2017scene}. We adopt UperNet framework~\cite{xiao2018unified} in the experiments. We train Upernet 160K iterations with batch size set as 16. We report the results in Table~\ref{tab:downstream} (a). The evaluation metric is mean Intersection of Union (mIoU) averaged over all semantic categories and we report single-scale test results here.
We compare our method with supervised pre-training on ImageNet-1K as well as state-of-the-art self-supervised models.
We can see that the proposed BootMAE gets superior performance than all the other baselines, further validating the effectiveness of our framework.

\begin{table*}[t]
\centering
%\tablestyle{6pt}{1.05}
\caption{(a) Semantic segmentation mIoU (\%) comparison on ADE20K. (b) Object detection and instance segmentation comparison in terms of box AP ($\text{AP}^{\text{bb}}$) and mask AP ($\text{AP}^{\text{mk}}$) on COCO. The same ViT-B backbone is used.}
\label{tab:downstream}
\setlength{\tabcolsep}{1mm}{
	%\scriptsize
\begin{tabular}{lcc||lccc}
\toprule
\multicolumn{3}{c||}{(a) Semantic segmentation} & \multicolumn{4}{c}{(b)Object detection and instance segmentation} \\
\hline
\multirow{2}{*}{ Models} & Pre-train  &  ADE-$20K$ &  \multirow{2}{*}{ Models}   & \qquad Pre-train \qquad  &  \multicolumn{2}{c}{COCO}\\  & epochs  & mIoU &  & \qquad epochs \qquad & $\text{AP}^{\text{bb}}$ & $\text{AP}^{\text{mk}}$\\

\hline
Supervised  & 300 &  47.4 & Supervised  &
 300 &  44.1 &  39.8 \\

MoCo~\cite{chen2021mocov3}          & 300 &  47.3 & 
MoCo~\cite{chen2021mocov3}          & 300 & 44.9  & 40.4  \\

BEiT~\cite{bao2021beit}           & 800 & 47.1 & 
BEiT~\cite{bao2021beit}          & 800 & 46.3  &  41.1  \\

MAE~\cite{he2021masked}         & 800 &  47.6 &
MAE~\cite{he2021masked}         & 800 & 46.8 & 41.9 \\

MAE$^*$~\cite{he2021masked}         & 1600 & 48.1  & 
MAE$^*$~\cite{he2021masked}          & 1600 & 47.2 & 42.0\\
\rowcolor{Graylight} 
BootMAE   & 800 & \textbf{49.1} & 
BootMAE   & 800 & \textbf{48.5} & \textbf{43.4} \\
\bottomrule
\end{tabular}}

\vspace{-2mm}
\end{table*}

\noindent \textbf{Object detection and segmentation.} 
We perform fine-tuning on the COCO object detection and segmentation~\cite{lin2014microsoftcoco}. We choose the Mask R-CNN~\cite{he2017mask} framework. Concretely, we adopt FPNs~\cite{lin2017feature} to scale the feature map into different size as introduced in~\cite{li2021benchmarking}.  The  fine-tuning  is  conducted  with  “1x”  (12  training epochs) schedule and single-scale input on the COCO training set.
The performance is tested on COCO validation set, following the strategy used in previous works~\cite{liu2021swin,dong2021cswin}. The results are reported in Table~\ref{tab:downstream} (b) in terms of box AP for detection and mask AP for segmentation. We observe that our model achieves 48.5\% for object detection and 43.4\% for segmentation, surpassing MAE by 1.3\% and 1.4\% respectively. 

\vspace{-1mm}
\section{Discussion and Conclusion}
\vspace{-1mm}
Masked image modeling for vision BERT pre-training has recently gain enormous popularity since the success of its counterpart, masked language modeling, in NLP domain. Recent tremendous progress of MIM has shown that masked signal modeling (including MIM and MLM) may potentially unify the self-supervised pre-training for both vision and language.
However, unlike language that has settled to an acknowledged framework, the research about MIM in vision is far from conclusive. There exists a lot of research exploring various directions considering the apparently distinct difference between language and vision.

In this paper, we introduce a new framework BootMAE with two core designs. 
(1) We propose to bootstrap the latent feature representation in MAE for better performance since the prediction target evolves with training, providing progressively richer information.
(2) We propose to decouple the target-specific context from the encoder so that the encoder focuses on modeling the image structure.
We present extensive experiments on various downstream tasks and comprehensive ablation studies to validate the effectiveness of the proposed framework.
In addition, we find that different prediction targets likely favor different masking strategies.

Previous MIM methods couple the target-specific information with structure learning in a single model. We argue that as the goal of MIM through inpainting is essentially modeling the within image structure, 
it is advantageous to enable the whole encoder to focus on semantic modeling which we empirically demonstrate its advantage in the experiments. In the future, we would like to seek theoretical connection between semantic modeling and representation learning.

\section{Acknowledgement.}
\noindent This work was supported in part by the Natural Science Foundation of China
under Grant U20B2047, 62072421, 62002334, and 62121002, Exploration Fund Project of University of Science and Technology of China under Grant YD3480002001, and by Fundamental Research Funds for the Central Universities under Grant WK2100000011.

\clearpage
\bibliographystyle{splncs04}
\bibliography{egbib}

\newpage
\section*{Appendix }
\appendix
\section{Cross attention in Regressor/Predictor }
One core design of our BootMAE is decoupling the target-specific context from the encoder, 
\emph{i.e.} provide low-level feature for the pixel regressor and high-level feature for the feature predictor. The feature injection procedure is conducted by a cross-attention operation. Formally, the cross-attention operator of regressor can be formulated as follows:
\begin{equation}
\begin{aligned}
    {\rm Cross Attention}&(Q,K,V) = {\rm softmax}(\frac{QK^T}{\sqrt{d_{decoder}}})V \\
    Q &= \bm{Z}_v W_Q \\ 
    K &= \bm{Z}_v^{shallow} W_K \\
    V &= \bm{Z}_v^{shallow} W_V 
\end{aligned}
\end{equation}
\begin{equation}
\begin{aligned}
    \bm{Z}_v^{'} = \bm{Z}_v + {\rm Cross Attention}(Q,K,V) 
\end{aligned}
\end{equation}
here $W_Q \in \mathbb{R}^{d_{decoder} \times d_{decoder} }$ project the input feature $\bm{Z}_v$ into the queue $Q$ with dimension $d_{decoder}$, and $W_Q \in \mathbb{R}^{d_{encoder} \times d_{decoder} }$ and $W_Q \in \mathbb{R}^{d_{encoder} \times d_{decoder} }$  project the injection feature $\bm{Z}_v^{shallow}$ into the key $K$ and value $V$. The $\bm{Z}_v^{shallow}$ can be replaced with $\bm{Z}_v^{deep}$ for the formulation of cross-attention in predictor.
With such cross-attention, we provide context information to decoder and relieve the encoder from ``memorizing" such context.

In our experiment, we use the output feature of the first encoder block as the low-level feature and the output feature after the last encoder block as the high-level feature. Such a simple strategy could be applied to models with different number of layers directly. Here we conduct some ablation to study how the injected feature affects the pretraining performance. 

As shown in Table \ref{tab:cross_feature}, for ViT-B with 12 blocks, we provide low-level feature (output of the first block), middle-level feature (output of the $6_{th}$ block), and high-level feature (output of the last block) to the pixel regressor and feature predictor respectively. We find that the pixel regression branch is sensitive to the level of injection feature. If we provide high-level or middle-level features to it, it performs poorly. Such a result proves our hypothesis that low-level context is crucial for the encoder to alleviate it from memorizing low-level context. If we provide high-level features to the pixel regressor, it is helpless and the encoder has to memorize the context anyway. On the contrary, we observe that the feature prediction is less sensitive to the provided context. One possible explanation is that the decoder input $\bm{Z}_v$ and the prediction target are both high-level semantic features, so the context information may be not so crucial. 

\begin{table}[t]
\centering%\scriptsize
%\tablestyle{6.5pt}{1.65}
\caption{Illustrating the effect of different injection feature. We show fine-tune accuracy (\%) on ImageNet-$1$K.}
\setlength{\tabcolsep}{2mm}{
\begin{tabular}{c|c| ccc}
\hline
 && \multicolumn{3}{c}{Context for Feature Predict} \\
 & & Low-level & Middle-level & High-level \\
\hline
\multirow{3}{*}{Context for Pixel Regress}& High-level & 83.2 & 83.3 & 83.4  \\ 
 & Middle-level & 83.6 & 83.6 & 83.7 \\
 & Low-level & 83.8 & 83.9 & \textbf{84.0} \\

\hline
\end{tabular}}
\vspace{-3mm}
\label{tab:cross_feature}
\end{table}

\section{Simple Feature Prediction for MAE}
As we mentioned in our introduction, we find simply replacing the pixel prediction task with feature prediction could help the MAE to get better performance. With a block-wise mask, MAE gets 83.8\% accuracy with only 800 epochs, even better than the vanilla MAE pretrained for 1600 epochs. Note that here we provide the prediction target feature from a 800 epochs pretrained MAE (row 2 in Table \ref{tab:mask_feat}).

\begin{table}[t]
	\centering
	\footnotesize
	\caption{Results of bootstrapped feature prediction. The performance is improved from 83.2\% to 83.6\% with the MAE features as prediction targets for 300 epochs, achieving the same performance with the vanilla MAE with pre-trained 1600 epochs. When the model is pre-trained for 800 epochs, it achieves 83.8\%.}
	\label{tab:mask_feat}
	\setlength{\tabcolsep}{4mm}{
		\begin{tabular}{l|cc|c }
			\hline
			Model & Prediction Target & Pre-train Epoch & Fine-tuning \\% & Linear probing  \\
			\hline
			MAE & Pixel & 300 & 83.2  \\
			MAE & Pixel & 800 & 83.4  \\
			MAE & Pixel & 1600 & 83.6  \\
			\hline
			MAE & Feature & 300 & 83.6 \\
			MAE & Feature & 800 & 83.8 \\
			%w/o feature prediction  &  83.3 &  \\
			%w/o pixel regressor  &  83.4 &    \\
			%\rowcolor{Graylight} 
			%Latent-MAE & 84.0 & \\
			\hline
	\end{tabular}}
	%\vspace{-3mm}
	
	\vspace{-4mm}
\end{table}

\section{Experiment Details}
In this section, we provide more detailed experimental settings.

\noindent \textbf{ImageNet Pretraining.}
We train our BootMAE with both ViT-B (12 transformer blocks with dimension 768) and ViT-L (24 transformer blocks with dimension 1024) for the encoder. 
The regressor and the predictor consist of 2 transformer blocks. The dimension of the regressor is set to 512 while the the dimension of the predictor is set to the same as the encoder for feature prediction. The learn-able mask token for regressor and the predictor are both initialized by random noise. 
The input is partitioned $14\times 14$ patches from the image of $224\times 224$, and each patch is of size 16x16. Following the setting in MAE, we only use standard random cropping and horizontal flipping for data augmentation.
The total masking ratio is 75\%, same with that in MAE~\cite{he2021masked}. The block-wise mask is generated follow the BEiT and we set the minima number for each block is 16 and the maximum is 60. 
Both ViT-B and ViT-L model are trained for 800 epochs with batch size set to 4096 and the learning rate is set to $1.5e^{-4}*batchsize/256$.
We use Adam~\cite{kingma2014adam} and a cosine schedule~\cite{loshchilov2016sgdr} with a single cycle and we warm up the learning rate for 40 epochs. The learning rate is further annealed following the cosine schedule. 
For the momentum parameter, we increase it from 0.999 to 0.9999 linearly in the first 100 epochs. For ViT-B, we further increase it to 0.99999 in the first 400 epochs.
We also use a weighted mask to assign larger loss weight to the center region of each block. 

\noindent \textbf{ADE20K Semantic segmentation.}
Here we use: UperNet~\cite{xiao2018unified} based on the implementation from mmsegmentaion~\cite{mmseg2020}. 
For UperNet, we follow the settings in ~\cite{bao2021beit} and use AdamW~\cite{loshchilov2017decoupled} optimizer with initial learning rate $4e^{-4}$, weight decay of 0.05 and batch size of 16 (8 GPUs with 2 images per GPU) for 160K iterations. The learning rate warmups with 1500 iterations at the beginning and decays with a linear decay strategy. We use the layer decay ~\cite{bao2021beit} for the backbone and we set it as 0.65. 
As the ViT architecture outputs features with the same size, here we add four different scale FPNs to scale the feature map into different size. Specifically, we upsample the output feature of the $4th$ block $4\times$, upsample the output feature of the $6th$ block $2\times$, keep the output feature of the $8th$ block unchanged and downsample the output feature of the $12th$ block $2\times$. 
We use the default augmentation setting in mmsegmentation including random horizontal flipping, random re-scaling (ratio range [0.5, 2.0]) and random photo-metric distortion. All the models are trained with input size $512\times512$. The stochastic depth is set to 0.1. When it comes to testing, we report single-scale test result.

\noindent \textbf{COCO Object Detection and Instance Segmentation.}
We use the classical object detection framework Mask R-CNN~\cite{he2017mask} based on the implementation from mmdetection~\cite{mmdetection}. We train the framework with $1\times$ schedule and single-scale input (image is resized so that the shorter side is 800 pixels, while the longer side does not exceed 1333 pixels) for 12 epochs. We use AdamW~\cite{loshchilov2017decoupled} optimizer with a learning rate of $4e^{-4}$, weight decay of 0.05 and batch size of 16. We also use the layer decay ~\cite{bao2021beit} for the backbone and we set it to 0.75. 
The learning rate declines at the $8th$ and $11th$ epoch with decay rate being 0.1. The stochastic depth is set to 0.1. 
Similar to the implementation of semantic segmentation above, we also use four different scale FPNs to scale the feature map into different size.

\end{document}